\def\year{2020}\relax
\documentclass[letterpaper]{article} % DO NOT CHANGE THIS
\usepackage{aaai20}  % DO NOT CHANGE THIS
\usepackage{times}  % DO NOT CHANGE THIS
\usepackage{helvet} % DO NOT CHANGE THIS
\usepackage{courier}  % DO NOT CHANGE THIS
\usepackage{url} % DO NOT CHANGE THIS
\usepackage{graphicx}  % DO NOT CHANGE THIS
\usepackage{amsmath}
\usepackage{amssymb}
\usepackage{multirow}
\newtheorem{thm}{Theorem}
\newtheorem{lem}{Lemma}
\usepackage{algorithm}
\usepackage[]{algpseudocode}

\usepackage{tikz}
\usetikzlibrary{shapes, arrows, positioning, decorations.markings}
\usetikzlibrary{calc} 
\definecolor {processblue}{cmyk}{0.96,0,0,0}
\tikzstyle{int}=[draw, fill=blue!20, minimum size=2em]
\tikzstyle{init} = [pin edge={to-,thin,black}]
\usetikzlibrary{shapes}
\usetikzlibrary{fit}
\usetikzlibrary{chains}
\usetikzlibrary{arrows}

\tikzset{
semi/.style={
  semicircle, ,top color =white , bottom color = processblue!20 ,
draw, processblue , text=blue,
  draw,
  minimum size=0.3cm
  }
}

\title{Aggregated Learning: A Vector-Quantization Approach to Learning Neural Network Classifiers}

\author{Masoumeh Soflaei,\textsuperscript{\rm 1} Hongyu Guo,\textsuperscript{\rm 2} Ali Al-Bashabsheh,\textsuperscript{\rm 3} Yongyi Mao,\textsuperscript{\rm 1} Richong Zhang\textsuperscript{\rm 3}
\\ \textsuperscript{\rm 1} University of Ottawa, Ottawa, Canada,
\textsuperscript{\rm 2}
National Research Council Canada,\\
\textsuperscript{\rm 3}
Beijing Advanced Institution on Big Data and Brain Computing, Beihang University, Beijing, China\\
msofl083@uottawa.ca, hongyu.guo@nrc-cnrc.gc.ca, entropyali@gmail.com,\\ yymao@eecs.uottawa.ca, zhangrc@act.buaa.edu.cn}

\begin{document}

\maketitle

\begin{abstract}
We consider the problem of learning a neural network classifier.
Under the information bottleneck (IB) principle, we associate with this classification problem a representation learning problem, which we call ``IB learning''. 
We show that IB learning is, in fact, equivalent to a special class of the quantization problem. The classical results in rate-distortion theory then suggest that IB learning can benefit from a ``vector quantization'' approach, namely, 
simultaneously learning the representations of multiple input objects. Such an approach assisted with some variational techniques, result in a novel learning framework, ``Aggregated Learning'',  for classification with neural network models. In this framework, several objects are jointly classified by a single neural network. The effectiveness of this framework is verified through extensive experiments on standard image recognition and text classification tasks.

\end{abstract}

\section{Introduction}

The revival of neural networks in the paradigm of deep learning \cite{LeCunBH15:DeepLearning} has stimulated intense interest in understanding the networking of deep neural networks, e.g., \cite{Tishby17:IB-deepLearning,ZhangBHRV16:rethinkingGeneralization}. Among various efforts, an information-theoretic approach, information bottleneck (IB) \cite{Tishby99} stands out as a fundamental tool to theorize the learning of deep neural networks \cite{Tishby17:IB-deepLearning,michael2018on,DBLP:conf/icml/DaiZGW18}.

Under the IB principle, the core of learning a neural network classifier is to find a representation $T$ of the input example $X$, that contains as little information as possible about $X$ and as much information as possible about the label $Y$. The conflict between these two requirements can be formulated as a constrained optimization problem in which one requirement is implemented as the objective function and another requirement as the constraint~\cite{Tishby03:IBratedistortion,Tishby99,Tishby10:IB-LearningGeneralization}. In this paper, we call this problem {\it IB learning}. 

A key observation that has inspired this work is that the optimization formulation of IB learning resembles greatly the rate-distortion function in rate-distortion theory, i.e., the theory for quantizing signals~\cite{Shannon59:rateDistortion}. A careful investigation along this direction indeed reveals that, conceptually, there is an unconventional quantization problem that is closely related to IB learning.  To that end, we formulate this problem, which we refer to as {\it IB quantization}. We prove that the objective of IB quantization, namely, designing quantizers that achieve the rate-distortion limit, is equivalent to the objective of IB learning. This result establishes an equivalence between the two problems.

In rate-distortion theory, it is well known that scalar quantizers, which quantize signals one at a time, are in general inferior to vector quantizers,
which quantize multiple signals at once.
The discovered equivalence between IB learning and IB quantization then suggests that IB learning may benefit from a ``vector quantization'' approach, in which the representations of multiple inputs are learned jointly.  Exploiting variational techniques and the recently proposed mutual information neural estimation (MINE) method \cite{belghazi2018mine}, we show that such a vector quantization approach to IB learning naturally results in a novel framework for learning neural network classifiers. We call this framework  {\it Aggregated Learning (AgrLearn)}.  

Briefly, in AgrLearn, $n$ random training objects are aggregated into a single amalgamated object and passed to the model; the model predicts the soft labels for all $n$ examples jointly. The training of an AgrLearn model is carried out by solving a min-max optimization problem, derived a variational relaxation of the IB learning problem and a MINE approximation of mutual information.

We conducted extensive experiments, applying AgrLearn to the current art of deep learning architectures for image and text classification. Our experimental results suggest that AgrLearn brings significant gain in classification accuracy. 
In practice, AgrLearn can be easily integrated into existing neural network architectures \footnote{Our implementation of AgrLearn is available at 
https://github.com/SITE5039/AgrLearn}. The proofs of theoretical results are provided in Appendices section.
%omitted due to length constraints. They will be included in an extended version of this paper.

\section{Information Bottleneck Learning}

The overall context of this work is a classification setting, where we let $\mathcal{X}$ denote the space of objects to be classified and $\mathcal{Y}$ denote the space of class labels.  Assume that the objects and labels are distributed according to an unknown distribution $p_{XY}$ on
$\mathcal{X}\times \mathcal{Y}$, where instead we are given a set
$\mathcal{D} := \{(X_1, Y_1), \dots, (X_N,Y_N)\}$ of i.i.d samples from $p_{XY}$. The objective of learning here is to find a classifier from ${\cal D}$ that classifies $X$ into its label $Y$.

Central to this classification problem is arguably the following representation learning problem: Find a representation of $X$ that only contains the information about $X$ relevant to its class label $Y$. Such a problem can be naturally formulated using the information bottleneck principle~\cite{Tishby99} and will be referred to as the {\em Information Bottleneck (IB) learning} problem. 

In IB learning, one is interested in learning a representation $T$ of $X$ 
in some space $\mathcal{T}$ such that the mutual information $I(X; T)$ between $X$ and $T$ is as small as possible whereas the mutual information $I(Y; T)$ between $T$ and the class label $Y$ is as large as possible. Such a representation is sensible since it aims at squeezing away all information in $X$ that is irrelevant to the classification task while keeping the relevant information intact. Intuitively, minimizing $I(X; T)$ forces the model not to over-fit to the irrelevant features of $X$, whereas maximizing $I(Y; T)$ extracts all features useful for the classification task. The two optimization objectives are in conflict with each other. A natural formulation to the IB learning problem is to consider one objective as the optimization objective and the other as a constraint.  
This gives rise to the following constrained optimization problem, subject to the Markov chain
$Y$---$X$---$T$, find 
\begin{equation}
\widehat{p}_{T|X} = \arg  \min _{p_{T|X}:I(X;T)\leq A} -I(Y;T), 
 \label{eq:IB2}
 \end{equation}
 for a nonnegative value $A$, or equivalently,
\begin{equation}
\widehat{p}_{T|X} = \arg  \min _{p_{T|X}:I(Y;T)\geq A'} I(X;T), 
 \label{eq:IB1}
   \end{equation}
for a nonnegative value $A'$. 
The Markov chain assumption ensures that any information in feature $T$ about label $Y$ is obtained from $X$ only.
For later use, we denote the minimum mutual information in \eqref{eq:IB1} as $R_{\rm IBL}(A')$, i.e.,
\begin{equation}
R_{\rm IBL}(A') =  \min _{p_{T|X}:I(Y;T)\geq A'} I(X;T). 
 \label{eq:RIB}
 \end{equation}
 
We note that solving this IB learning problem, i.e., obtaining the optimal 
$\widehat{p}_{T|X}$ and its corresponding bottleneck representation $T$ does not automatically solve the classification problem. It is still required to build a classifier that predicts the class label $Y$ based on the representation $T$ of $X$. Nonetheless later in this paper, we will show that solving a variational approximation of the IB learning problem may, in fact, provide a direct solution to the classification problem of interest.

\section{Information Bottleneck Quantization}

We now formulate the {\em Information Bottleneck (IB) quantization problem}.  Our objective in this section is to show that the IB quantization and IB learning problems are equivalent.

Let $(X_1,Y_1),(X_2,Y_2),\dots,(X_n,Y_n)$ be drawn i.i.d from $p_{XY}$. 
The sequences $(X_1,X_2,\cdots, X_n)$ and $(Y_1,Y_2,\cdots, Y_n)$ are denoted by $X^n$ and $Y^n$, respectively. 

An $(n,2^{nR})$ {\it IB-quantization code} is a pair $(f_n,g_n)$ in which $f_n$ maps each sequence $X^n$ to an integer in $\lbrace 1, 2, \cdots, 2^{nR} \rbrace$ and $g_n$ maps an integer in $\lbrace 1, 2, \cdots, 2^{nR} \rbrace$ to a sequence $T^n:=(T_1,T_2,\cdots,T_n) \in {\mathcal T}^n$. 
Using the standard nomenclature in quantization, the quantity $R$ is referred to as the {\it rate} of the code and $n$ as the {\it length} of the code. Using this code, $f_n$ encodes the sequence $X^n$ as the integer $f_n(X^n)$ and $g_n$ reconstructs $X^n$ as a representation $T^n := g_n(f_n(X^n))$.

Unlike standard quantization problems, the IB quantization problem uses a distortion measure that may depend on the code. To that end, 
 for any $x \in \mathcal{X}$, $t \in \mathcal{T}$ and any two conditional distributions $q_{Y|X}$ and $q_{Y|T}$, define 
\begin{equation}
 d_{\rm IB}(x,t;q_{Y|X},q_{Y|T}) := \text{KL}(q_{Y|X}(.|x)\Vert q_{Y|T}(.|t)),
 \end{equation}  
 where $\text{KL}(.\Vert .)$ is the Kullback--Leibler (KL) divergence.  

  Note that the code $(f_n,g_n)$, together with $p_{XY}$, induce a joint distribution over the Markov chain $Y^n$---$X^n$---$T^n $.
 Under this joint distribution the conditional distributions $p_{Y_i |X_i}$ and $p_{Y_i |T_i}$ are well
defined for each $i=1,2,...,n$. Hence, given the code $(f_n, g_n)$ and
for any two sequences $x^n\in \mathcal{X}^n$ and $t^n\in \mathcal{T}^n$, their {\it IB distortion} is defined as:
 \begin{equation}
\overline{d}_{\rm IB}(x^n,t^n) := \frac{1}{n}\sum_{i=1}^n d_{\rm IB}(x_i,t_i;p_{Y_i|X_i},p_{Y_i|T_i}), 
\end{equation} 
We note that the quantity $\overline{d}_{\rm IB}(x^n,t^n)$ measures a ``loss of information about $Y$'' when the code $(f_n, g_n)$ is used to represent $x^n$ as $t^n$. Specifically, consider the source coding problem of compressing $Y^n$ based on observing $X^n=x^n$. If the conditional distribution $p_{Y_i|X_i}(\cdot|x_i)$ for each $i$ is mistaken as $p_{Y_i|T_i}(\cdot|t_i)$ in the  design of the source code, the average additional coding overhead per $Y$-symbol is precisely $\overline{d}_{\rm IB}(x^n,t^n)$.

Using this distortion measure, the {\it IB quantization problem} is to find a code $(f_n,g_n)$ having
the smallest rate $R$ subject to the constraint $\mathbb{E}\overline{d}_{\rm IB}(X^n,T^n)\leq D$, where
$\mathbb{E}$ denotes expectation. For given $p_{XY}$ and $\mathcal{T}$, a rate
distortion pair $(R,D)$ is called {\em achievable} if $\mathbb{E}\overline{d}_{\rm IB}(X^n,T^n)\leq D$ for some
sequence of $(f_n,g_n)$ codes. 
As usual, the {\it rate-distortion} function for the IB quantization problem, which we denote by 
 $R_{\rm IBQ}(D)$,  is defined as the smallest rate $R$ such that $(R,D)$ is achievable.
 
 \begin{thm}
\label{th:rate-distortion}
Given $p_{XY}$ and $\mathcal{T}$, the rate-distortion function for the IB quantization problem can be written as 
 \begin{equation}
R_{\rm IBQ}(D) = \min_{p_{T|X}: \mathbb{E}d_{\rm IB}(X,T)\leq D}  I(X;T) 
    \label{eq:rd}
\end{equation} 
where the expectation is defined as
\[
\mathbb{E} d_{\rm IB}(X,T)  :=\sum_{x,t} d_{\rm IB}(x,t; p_{Y|X}, p_{Y|T}) p_{XT}(x,t). 
\] 
\end{thm}

This theorem provides a limit on the 
achievable rates of the IB quantization problem.
We note that this result was first shown in \cite{Tishby03:IBratedistortion}. However in \cite{Tishby03:IBratedistortion},  the result relies on the assumption that $|\mathcal{T}| \geq |\mathcal{X}|+2$, whereas in this theorem the condition is removed. 

The form of the rate-distortion function $R_{\rm IBQ}$ for the IB quantization problem given in
Theorem \ref{th:rate-distortion} resembles greatly the optimal objective of IB learning $R_{\rm IBL}$ in \eqref{eq:RIB}.
More precisely, we have

\begin{thm} 
$R_{\rm IBL}(A')= R_{\rm IBQ}(I(X;Y)-A')$
\end{thm}

\noindent {\em Proof:} 
We have
\begin{eqnarray*}
\mathbb{E} d_{\rm IB}(X,T) & :=&\sum_{x,t} d_{\rm IB}(x,t; p_{Y|X}, p_{Y|T}) p_{XT}(x,t)  \\
& =& I(X;Y) - I(Y;T)
%\label{eq:dist-expect3}
\end{eqnarray*}
where the second equality is by the definition of $d_{\rm IB}$ and the Markov chain $Y$---$X$---$T$ assumption. 
Hence, we may rewrite (\ref{eq:rd}) in Theorem \ref{th:rate-distortion} as
\begin{eqnarray*}
	R_{\rm IBQ}(D) 
	& = & \min_{p_{T|X}:
	I(X;Y) - I(Y;T)\leq D} I(X;T) \\
	& = & \min_{p_{T|X}: I(Y;T) \geq I(X;Y) -  D} I(X;T) \\
	& = & R_{\rm IBL}(I(X;Y) - D)
\end{eqnarray*}
The theorem follows by substituting $A':=I(X;Y) - D$. \hfill $\Box$
 
This theorem relates the IB learning and IB quantization problems, where we note that $I(X; Y)$ is a constant that only depends on $p_{XY}$. By this theorem, solving the IB learning problem where the information about $Y$ contained in $T$ needs to be no less than $A'$ is equivalent to solving the IB quantization problem so that the distortion is no more than $I(X; Y) - A'$.

\section{Variational Approach to IB Learning}

Having established the equivalence between IB learning and IB quantization, we now turn to solve
the IB learning problem. The objective of this section is to develop a variational approach to this
problem which not only provides a bottleneck representation $T$ for $X$ but also leads to a
classifier for the classification problem at hand. We note that the results presented in this section
also underlies the ``variational information bottleneck'' approach of \cite{DVIB:AlemiFD016}. 

 We first establish the following result.
 
\begin{thm} 
\label{th:varitionalbound} 
Under any distribution $p_{YXT}$ that satisfies the Markov chain $Y$---$X$---$T$, we have 
\begin{equation}
	I(Y; T) \ge \mathbb{E}_{(x,y)\sim p_{XY}, \atop{t\sim p_{T|X}(\cdot|x)} }
\log q_{Y|T}(y|t)
 +H(Y)
\end{equation}
for any conditional distribution $q_{Y|T}$ of a random variable on ${\cal Y}$ conditioned on $T$.
In addition, the above inequality holds with equality if and only if $q_{Y|T}$ is equal to $p_{Y|T}$. 
\end{thm}

As a consequence of this theorem, the mutual information $I(Y;T)$ can be written as 
\begin{eqnarray*}
	I(Y; T) = \max_{q_{Y|T}}& \mathbb{E}_{(x,y)\sim p_{XY}, \atop{t\sim p_{T|X}(\cdot|x)} } \log q_{Y|T}(y|t)
	+H(Y).
\end{eqnarray*}
Substituting this in the IB learning problem as formulated in (\ref{eq:IB2}), we have
\begin{align*}
	\widehat{p}_{T|X}
	&=
	\arg\min _{p_{T|X}:I(X;T)\leq A} -I(Y;T) \\
	&=
	\arg
	\min _{p_{T|X}:\atop{I(X;T)\leq A} }
	\left\{
	-
	%\underset{a}{}
	\max_{q_{Y|T}}
	\mathbb{E}_{
	\kern -.5em (x,y)\sim p_{XY},\atop{t\sim p_{T|X}(\cdot|x)}
}
	\log q_{Y|T}(y|t) 
	 \right\}
	 \\
	&=
	\arg\min
	%\argmin
	_{p_{T|X}:\atop{I(X;T)\leq A} }
	\min_{q_{Y|T}}
	\left\{
		-\mathbb{E}_{\kern-.5em (x,y)\sim p_{XY}, \atop{t\sim p_{T|X}(\cdot|x)} }  
	\log q_{Y|T}(y|t)
	\right\}
\end{align*}

Now suppose we have a neural network representing the mapping $p_{T|X}$ and that we represent
$q_{Y|T}$ using another network. Then we may construct an overall network by concatenating the two
networks. Specifically, each object $x$ will be first passed to the network  $p_{T|X}$, and the output $T$ of the network is passed to the network $q_{Y|T}$. If the true class label $y$ is modeled as being generated from this concatenated network, it is easy to see that the
cross-entropy loss $\ell_{\rm CE}$ of the network is the expectation above, i.e.,
\begin{align}
	\ell_{\rm CE} =  -\mathbb{E}_{(x,y)\sim p_{XY}, t\sim p_{T|X}(\cdot|x)}  \log q_{Y|T}(y|t).
	\label{eqq:lce}
\end{align}
In other words, the IB learning problem can be formulated as  solving the following optimization problem:
\begin{equation}
    \min_{p_{T|X}, q_{Y|T}} \ell_{\rm CE}\left(p_{T|X}, q_{Y|T}\right) ~{\rm subject~ to~} I(X;T)\leq A
\end{equation}
Hence, introducing a Lagrange multiplier, subsequently we will focus on the following unconstrained problem
\begin{equation}
\label{eq:variationForm}
    \min_{p_{T|X}, q_{Y|T}} \ell_{\rm CE}\left(p_{T|X}, q_{Y|T}\right) + \alpha I(X;T)
\end{equation}
for nonnegative $\alpha$.

An apparent advantage of this approach to IB learning is that when the optimization problem
(\ref{eq:variationForm}) is solved, not only is the bottleneck representation $T$ found, but also
the entire classification network is obtained. 

It is worth noting that the variational formulation (\ref{eq:variationForm}) of IB learning can be
viewed as a generalization of learning with standard neural networks under the cross-entropy loss.
Specifically,  learning with standard neural networks is a reduction of (\ref{eq:variationForm})
in which
the standard neural network contains no term $ \alpha I(X;T)$, or equivalently has
$\alpha=0$. 

The generalization of learning with standard neural networks to the formulation of IB learning in  (\ref{eq:variationForm}) is arguably beneficial in two respects: 

\begin{enumerate}
    \item The $\alpha I(X; T)$ regularization term in (\ref{eq:variationForm}) serves to control the model complexity so as to reduce the generalization gap.
    \item Generalizing the deterministic map from $X$ to $T$ in standard neural networks to a stochastic one in (\ref{eq:variationForm}) minimizes the cross-entropy loss $\ell_{\rm CE}$ over a larger space; this potentially allows further decrease of $\ell_{\rm CE}$, thereby achieving better classification accuracy. 
    We note that the ``Deep Variational Information Bottleneck'' (DVIB) approach of \cite{DVIB:AlemiFD016}, not necessarily motivated by the same reason, uses the same variational bound of $I(Y; T)$ and arrives at the same formulation as (\ref{eq:variationForm}). 
    
\end{enumerate}

In the remainder of this paper, we present a new strategy, termed ``Aggregated Learning'', to implement the IB learning formulation (\ref{eq:variationForm}).

\section{Aggregated Learning (AgrLearn)}

We now introduce the Aggregated Learning (AgrLearn) framework for learning with neural networks. 
 We will stay with the IB learning formulation of (\ref{eq:variationForm}) while keeping in mind that it results from a variational approximation of the formulation in (\ref{eq:IB2}).

Recall from Theorem \ref{th:rate-distortion} that the IB learning problem is equivalent to the IB quantization problem. In the classical rate-distortion theory \cite{Shannon59:rateDistortion}, it is well known that in order to achieve the rate-distortion limit of quantization, in general, one must consider the use of {\em vector quantizers}. 

In the context of 
IB quantization, a {\em vector quantizer} is an IB-quantization code $(f_n, g_n)$ with $n>1$ whereas a scalar quantizer is an IB-quantization code $(f_n, g_n)$ with $n=1$. From rate-distortion theory, better quantizers result from using quantization codes with larger length $n$. In particular, in order to achieve the rate-distortion function, it is in general required that the length $n$ of the rate-distortion code be made asymptotically large. 

Note that a scalar IB-quantization code $(f_1, g_1)$ maps $X$ to $T$ by
\[
T=g_1(f_1(X)):=(g_1\circ f_1)(X).
\]  
Under the equivalence
between IB quantization and IB learning, the mapping $g_1\circ f_1$ induced by the scalar quantizer $(f_1, g_1)$ essentially defines a  conditional distribution $p_{T|X}$ in IB learning, which simply reduces to the deterministic function $g_1\circ f_1$. On the other hand, in learning with a standard neural network, the deterministic mapping, say $h$, from the input space ${\cal X}$ to the bottleneck space ${\cal T}$ (which could 
refer to the space of feature representation at any intermediate layer of the network), can be regarded as implementing a scalar IB-quantization code $(f_1, g_1)$ with
\[
g_1\circ f_1=h.
\]

The superiority of vector quantizers to scalar quantizers then motivates us to develop a vector-quantization approach to IB learning, which we call Aggregated Learning or AgrLearn in short. -- Like a vector quantizer, which quantizes $n$ signals simultaneously, AgrLearn classifies $n$ input objects jointly at the same time, the details of which are given below. 

The framework of AgrLearn consists  of two networks, which we refer to as the ``main network'' and the ``regularizing network'' respectively. 

\subsection{The Main Network} 

The main network takes as its input the concatenation of $n$ objects $(X_1, X_2, \ldots, X_n):=X^n$. Such a concatenated input will be  referred to as an ``$n$-fold aggregated input''. 
 
\begin{figure}[t]
\scalebox{0.65}{
\begin{tikzpicture}[-latex ,auto ,node distance =2 cm and 2 cm ,on grid ,
semithick ,
state/.style ={ circle ,top color =white , bottom color = cyan!60 ,
draw, cyan , text=blue , minimum width =0.1cm},
box/.style ={rectangle ,top color =white , bottom color = cyan!60 ,
draw, cyan , text=blue , minimum width =1.7cm , minimum height = 0.5cm, rounded corners},
highbox/.style ={rectangle ,top color =white , bottom color = cyan!60 ,
draw, cyan , text=blue , minimum width =1.7cm , minimum height = 1.6cm, rounded corners},
neuron/.style ={rectangle ,top color =white , bottom color = red!20 ,
draw, red , text=red , minimum width =2.6cm , minimum height = 3.6cm, rounded corners},
triangle/.style = {top color =white , bottom color = cyan!60 ,
draw, cyan , text=blue, regular polygon, regular polygon sides=3, minimum size=0.5cm, draw },
node rotated/.style = {rotate=270},
    border rotated/.style = {shape border rotate=270}]
% \node[neuron]{};
\node[](realcenter){};
\node[](center)[above=0.2cm of realcenter]{};
\node[highbox](cnn)[above=0.5cm of center]{$h$};
\node[state](sold)[left=2.5cm of cnn]{};
\path (sold) edge [] node[]{$X^n$} (cnn);
\node[](lcenter0)[left=0.5cm of sold]{};
\node[](lcenter1)[above=1.5cm of lcenter0]{};
\node[](lcenter2)[above=0.75cm of lcenter0]{};
\node[](lcentern)[below=1.5cm of lcenter0]{};
\node[](lcenter11)[left=2.25cm of lcenter1]{$X_1$};
\node[](lcenter21)[left=2.25cm of lcenter2]{$X_2$};
\node[][left=2.25cm of lcenter0]{$\vdots$};
\node[](lcentern1)[left=2.25cm of lcentern]{$X_n$};
\draw [->] (lcenter11) to [out=0,in=180] (sold);
\draw [->] (lcenter21) to [out=0,in=180] (sold);
\draw [->] (lcentern1) to [out=0,in=180] (sold);
\node[highbox](snew)[right=3.5cm of cnn]{$q_{Y^n|T^n}$};
 \node[](rcenter0)[right=2.5cm of snew]{};
 \node[](rcenter1)[above=1.5cm of rcenter0]{};
  \node[](rcenter2)[above=0.75cm of rcenter0]{};
  \node[](rcentern)[below=1.5cm of rcenter0]{};
\node[](rcenter11)[right=0.5cm of rcenter1]{$q_{Y_1|X^n}$};
\node[](rcenter21)[right=0.5cm of rcenter2]{$q_{Y_2|X^n}$};
\node[](rcentern1)[right=0.5cm of rcentern]{$q_{Y_n|X^n}$};
\node[]()[right=0.5cm of rcenter0]{$\vdots$};
\path (cnn) edge [] node[]{$T^n$}  (snew);
\draw [->] (snew) to [out=10,in=180] (rcenter11);
\draw [->] (snew) to [out=0,in=180] (rcenter21);
\draw [->] (snew) to [out=-10,in=180] (rcentern1);
\end{tikzpicture}}
\caption{The main network in AgrLearn. The small circle denotes concatenation.}
\label{fig:Agrlearn}
\end{figure}
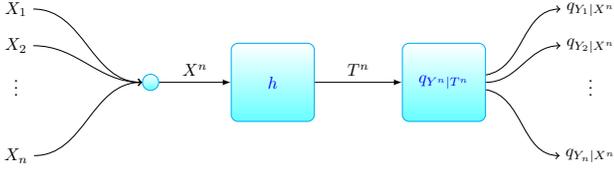

The main network consists of two parts, as seen in Figure \ref{fig:Agrlearn}. The first part, or the ``pre-bottleneck'' part, implements a deterministic mapping $h:{\cal X}^n \rightarrow {\cal T}^n$ that maps an aggregated input $X^n$ to an ``aggregated bottleneck'' $T^n$ via
\begin{equation}
T^n:= (T_1, T_2, \ldots, T_n):= h(X^n). 
\end{equation}
 The second part, or the ``post-bottleneck'' part, implements a stochastic mapping $q_{Y^n|T^n}$ from ${\cal T}^n$ to ${\cal Y}^n$ that factorizes according to 
 \begin{equation}
 \label{eq:postbot}
     q_{Y^n|T^n}(y^n|t^n) : = \prod_{i=1}^n q_{Y_i|T^n}(y_i|t^n)\\
 \end{equation}
 Overall the main network expresses a stochastic mapping from ${\cal X}^n$ to ${\cal Y}^n$, which can be expressed as 
 \begin{equation}
 \label{eq:agrLearnNet_main}
     q_{Y^n|X^n}(y^n|x^n): = 
     \prod_{i=1}^n q_{Y_i|T^n}(y_i|h(x^n))\\
 \end{equation}
On the main
 network as specified by 
 (\ref{eq:agrLearnNet_main}), 
 define
\begin{equation}
\label{eq:agrLearn_CEloss}
\ell_{\rm CE}^{(n)} : =  -\mathbb{E}_{x^ny^n\sim p_{XY}^{\otimes n}}  
\log q_{Y^n|X^n}(y^n|x^n)
\end{equation}
where $p_{XY}^{\otimes n}$ is the distribution on $\left({\cal X}\times {\cal Y}\right)^n$ induced by drawing $n$ samples i.i.d. from $p_{XY}$. Clearly $\ell_{\rm CE}^{(n)}$ is nothing more than the cross-entropy loss of the network's predictive distribution $q_{Y^n|X^n}$  for the aggregated
input $X^n$ with respect to their labels $Y^n$.  As we will be minimizing this cross-entropy loss function, we next discuss its properties. 

Following Theorem \ref{th:varitionalbound}, 
\begin{equation}
\ell^{(n)}_{\rm CE} \ge  nH(Y)   -I(Y^n; T^n). 
\end{equation}
and if the post-bottleneck network component $q_{Y^n|T^n}$ has sufficient capacity, then
\[
\min_{q_{Y^n|T^n}} 
\ell_{\rm CE}^{(n)} 
= n H(Y) -  I (Y^n; T^n)
\]

That is if the post-bottleneck component has sufficient capacity, then minimizing $\ell_{\rm CE}^{(n)}$ over the entire main network also maximizes $I(Y^n; T^n)$.

\subsection{The Regularizing Network}
 
The regularizing network is essentially a mutual information neural estimator (MINE) network \cite{belghazi2018mine}, which serves to estimate $I(X; T)$ and penalizes it during the training of the main network.  For a careful development of MINE, the reader is referred to \cite{belghazi2018mine}. Here we only give a brief description. 

\noindent {\bf MINE in a Nutshell} Suppose that 
${\cal U}$ and ${\cal V}$ are two spaces and that 
there is a joint distribution $p_{UV}$ on ${\cal U} \times {\cal V}$ defining a pair $(U, V)$ of random variables. Suppose that we can perform i.i.d. sampling  of $p_{UV}$
and we wish to estimate the mutual information $I(U; V)$ from the samples. 
In the framework of MINE, a family $\Gamma$ of functions is constructed as a neural network, where each 
$\gamma\in \Gamma$ is a function mapping ${\cal U}\times {\cal V}$ to the set ${\mathbb R}$ of real numbers.  Then due to dual representation of KL divergence~\cite{donsker1983asymptotic}, the mutual information $I(U; V)$ can be estimated as 

 \begin{align}
   \begin{split}
       \label{eq:MINE_theory}
     \widehat{I}(U;V) :=& \max_{\gamma \in \Gamma} \lbrace
    {\mathbb E}_{(u, v)\sim p_{UV}}
 \gamma(u, v) \\
    & - \log 
     {\mathbb E}_{(u, v)\sim p_U\otimes p_V}
      \exp
     \left( \gamma(u, v)
     \right)
     \rbrace
      \end{split}
      \end{align}
 We will denote the term that gets maximized in 
(\ref{eq:MINE_theory}) by $J(U, V; \gamma)$, namely,
 \begin{align}
   \begin{split}
J(U, V; \gamma): =& 
{\mathbb E}_{(u, v)\sim p_{UV}}
 \gamma(u, v)\\
    & - \log 
     {\mathbb E}_{(u, v)\sim p_U\otimes p_V}
      \exp
     \left( \gamma(u, v)
     \right)
 \end{split}
      \end{align}
      
and re-express $\widehat{I}(U; V)$ as
\[
\widehat{I}(U; V) = \max_{\gamma\in \Gamma} J(U, V; \gamma)
\]
As usual, practical computation of $J(U, V; \gamma)$ exploits Monte-Carlo approximation based on samples drawn from $p_{UV}$. 
A natural way to apply MINE to the estimation of $I(X; T)$ in AgrLearn is taking ${\cal U}:={\cal X}^n$, ${\cal V}:={\cal T}^n$, $U=X^n$, $V=T^n$.
  
  This allows us to estimate $I(X^n; T^n)$ by
\begin{equation}
\widehat{I}(X^n; T^n) = \max_{\gamma \in \Gamma} J(X^n, T^n; \gamma)
\end{equation}
where $T^n$ is computed by the pre-bottleneck component of the main network with $X^n$ as its input.  We may then take $\widehat{I}(X^n; T^n)$ as an approximation of $n I(X; T)$.  The network implementing the computation of $J(X^n, T^n; \gamma)$
is referred to as the regularizing network.
  \begin{algorithm} 
\caption{ Training in $n$-fold AgrLearn}
\label{algo:trainAgrLearn}
\begin{algorithmic}[0]
\State Initialize $h, q_{Y^n|T^n}$, and $\gamma$\;
\While{not stop training}
   \State Draw $m\times n$ examples to form a batch of $m$ $n$-fold aggregated examples  $\left\{x^n_{(1)}, x^n_{(2)}, \ldots, x^n_{(m)}\right\}$\;
    \For {$k=1$ \textbf{\rm to } $K$}
        \For { $i=1$ \textbf{\rm to} $m$}
          \State  $t^n_{(i)}:=h(x^n_{(i)})$
        \EndFor
        \State Select a random permutation $\tau$ on $\{1, 2, \ldots, m\}$\;
         \State Forward compute 
        $J:=\frac{1}{m}\sum\limits_{i=1}^m \gamma(x^n_{(i)}, t^n_{(i)}) - \phantom{mmmmmmmmmmm} \log \frac{1}{m}\sum\limits_{i=1}^m \exp \left( \gamma(x^n_{(i)}, t^n_{(\tau(i))}) \right)$\;
         \State  $\gamma \gets \gamma + \lambda_{\rm in} \cdot \frac{\partial J}{\partial \gamma}$\;
    \EndFor
    \State Select a random permutation $\tau$ on $\{1, 2, \ldots, m\}$\;
    \State Forward compute 
    $J:=\frac{1}{m}\sum\limits_{i=1}^m \gamma(x^n_{(i)}, t^n_{(i)}) - \phantom{mmmmmmmmmmm} \log \frac{1}{m}\sum\limits_{i=1}^m \exp \left( \gamma(x^n_{(i)}, t^n_{(\tau(i))}) \right)$\;
    \State Forward compute $\ell:= \frac{1}{m}\sum\limits_{i=1}^m \log q_{Y^n|T^n}(y^n_{(i)}|t^n_{(i)})$\;
    \State Compute $\Omega:= \ell + \alpha \cdot J$, \;
    %\State 
    $h \gets h - \lambda_{\rm out}\cdot \frac{\partial \Omega}{\partial h}$, and\;
    \State $q_{Y^n|T^n} \gets q_{Y^n|T^n} - \lambda_{\rm out}\cdot \frac{\partial \Omega}{\partial q_{Y^n|T^n}}$\;
\EndWhile
\end{algorithmic}
\end{algorithm}

 \subsection{Training and Prediction}

With this development, we may define an overall objective function $\Omega (h, q_{Y^n|T^n}, \gamma)$ as
\begin{equation}
    \Omega (h, q_{Y^n|T^n}, \gamma):= 
    \ell_{\rm CE}^{(n)} + \alpha J(X^n, T^n; \gamma)
    \label{eq:loss}
\end{equation}
where we note that the term $\alpha J(X^n, T^n; \gamma)$ also depends on $h$ implicitly. The above development then suggests that solving the IB learning problem in the form of (\ref{eq:variationForm}) can be approximated by solving the following min-max problem:

\begin{equation}
\label{eq:minMax}
\min_{h, q_{Y^n|T^n}} \max_{\gamma} 
\Omega (h, q_{Y^n|T^n}, \gamma)
\end{equation}

In the training of AgrLearn, mini-batched SGD can be used to solve the above min-max problem. The training algorithm is given in Algorithm \ref{algo:trainAgrLearn}.

In the prediction phase, ``Replicated Classification" protocol is used\footnote{Two additional protocols were also investigated. {\it Contextual Classification}: For each object $X$, $n-1$ random examples are drawn from the training set $\mathcal{D_\mathcal{X}}$ and concatenated with $X$ to form the input; the predictive distribution for $X$ generated by the model is then retrieved. This process is repeated $k$ times, and the average of the $k$ predictive distribution is taken as the label predictive distribution for $X$. {\it Batched Classification}: Let $\mathcal{D^{\text{test}}_\mathcal{X}}$ denote the set of all objects to be classified. In Batched Classification, $\mathcal{D^{\text{test}}_\mathcal{X}}$ are classified jointly through drawing $k$ random batches of $n$ objects from $\mathcal{D^{\text{test}}_\mathcal{X}}$. The objects in the $i^{th}$ batch $B_i$ are concatenated to form the input and passed to the model. The final label predictive distribution for each object $X$ in $\mathcal{D^{\text{test}}_\mathcal{X}}$ is taken as the average of the predictive distributions of $X$ output by the model for all batches $B_i$'s containing $X$. Since we observe that all three protocols result in comparable performances, all results reported in the paper are obtained using the Replicated Classification protocol.}. Each object $X$ is replicated $n$ times and concatenated to form the input. The average of $n$ predictive distributions generated by the model is taken as the label predictive distribution for $X$.

\section{Experimental Studies}

We evaluate AgrLearn with deep network architectures such as ResNet for classification tasks in both image
and natural language domains. Standard benchmarking datasets are used. 

We use mini-batched backprop for 400 epochs\footnote{Here an epoch refers to going over $N$ aggregated training examples, where $N = |\mathcal{D}_\mathcal{X}|$.} with exactly the same hyper-parameter settings without dropout. Specifically, weight decay is $10^{-4}$, and each mini-batch contains 64 aggregated training examples. The learning rate for the main network is set to 0.1 initially and decays by a factor of $10$ after $100$, $150$, and $250$ epochs. 
Each reported performance value (error rate or accuracy) is the median of the performance values obtained in the final 10 epochs by averaging that value over running the same setting 7 times. 

\subsection{Image Recognition}
Experiments are conducted on the \textbf{CIFAR-10}, \textbf{CIFAR-100} datasets with two widely used deep network architectures, namely ResNet~\cite{he2016identity} and WideResNet~\cite{zagoruyko2016wide}. The \textbf{CIFAR-10} dataset has 50,000 training images, 10,000 test images, and 10 image classes, and the \textbf{CIFAR-100} dataset is similar to CIFAR-10 but with 100 classes.

We apply AgrLearn to the 18-layer and 34-layer Pre-activation ResNet (\textit{ResNet-18} and \textit{ResNet-34})~\cite{he2016identity} as implemented in~\cite{liu17}, and the 22-layer WideResNet (\textit{WideResNet-22-10})~\cite{zagoruyko2016wide} as implemented in~\cite{Zagoruyko/code}. The resulting AgrLearn model differs from original ResNet and WideResNet in its $n$ parallel soft-max layers in post-bottleneck part(as opposed to the single soft-max layer in ResNet and WideResNet) and the number of filters in the last layer of pre-bottleneck part, which is expanded by factor $n$. This expanding by factor $n$ is required because the input dimension in AgrLearn increases significantly, and the model is required to extract joint features across individual objects in the amalgamated example. 

Note that fold number $1$ (fold-1) denotes the standard neural network in which just one object passes to the network and fold number greater than $1$ denotes an AgrLearn framework wherein
multiple objects are aggregated and passed to the network. The quantity $\alpha$ is the coefficient of the second term in
(\ref{eq:loss}), in which $\alpha=0$ corresponds to that only the cross-entropy loss is considered
, and $\alpha>0$ corresponds to that the regularization network is added to the main network.

\subsubsection{Predictive Performance}
The prediction error rates of AgrLearn for different number of folds are shown in Tables \ref{tab:res18}, \ref{tab:wideres}, and  \ref{tab:res34}. 

It can be seen that AgrLearn significantly boosts the performance of ResNet-18, ResNet-34 and WideResNet-22-10. For example, with respect to ResNet-18, the relative error reductions achieved by fold-2, where $\alpha=0$  
are $3.74$\%, and $2.83$\% on CIFAR-10, and CIFAR-100, and where $\alpha>0$ 
the reductions are $3.86$\%, and $3.21$\% on CIFAR-10, and CIFAR-100 respectively. 
 
Similarly significant improvement upon ResNet-34 and WideResNet is also observed. For example, with respect to WideResNet-22-10,  the relative error reductions achieved by fold-2, where $\alpha=0$, are $2.56$\%, and $3.93$\% on CIFAR-10, and CIFAR-100, and where $\alpha>0$, the reductions are $1.18$\%, and $3.89$\% on CIFAR-10, and CIFAR-100 respectively. The relative error reductions with respect to ResNet-34, achieved by fold-2, where $\alpha=0$ are $5.26$\%, and $5.16$\% on CIFAR-10, and CIFAR-100, and where $\alpha>0$, the reductions are $5.3$\%, and $6.59$\% on CIFAR-10, and CIFAR-100 respectively. 
\begin{table}[htb]
  \caption{\label{tab:res18} Test error rates (\%) of ResNet-18 and its AgrLearn counterparts on CIFAR-10, and CIFAR-100}
  \centering
  \begin{tabular}{c|cc|cc}
    \hline
     \multirow{2}{*}{\bf{Dataset}}  &\multicolumn{2}{c|}{\bf fold-1}       &\multicolumn{2}{c}{\bf fold-2}      \\  
       &$\alpha = 0$ &$\alpha = 0.7$    &$\alpha = 0$ &$\alpha = 0.3$       \\ \hline
    CIFAR-10          &   5.08  &  4.92  &  4.89   &  4.73      \\ 
    CIFAR-100        &	23.7    &   23.7    &  23.03     & 22.94  \\ \hline
    \end{tabular}
\end{table}
\begin{table}[htb]
  \caption{\label{tab:wideres} Test error rates (\%) of WideResNet-22-10 and its AgrLearn counterparts on CIFAR-10, and CIFAR-100}
  \centering
  \begin{tabular}{c|cc|cc}
    \hline
    \multirow{2}{*}{\bf{Dataset}}  &\multicolumn{2}{c|}{\bf fold-1}       &\multicolumn{2}{c}{\bf fold-2}      \\  
       &$\alpha = 0$ &$\alpha = 0.7$    &$\alpha = 0$ &$\alpha = 0.3$       \\ \hline
    CIFAR-10          &  4.3          & 4.23   & 4.19       & 4.18      \\ 
    CIFAR-100        &	21.13    &   21.1   &   20.3    & 20.28 \\ \hline
    \end{tabular}
\end{table}
\begin{table}[htb]
  \caption{\label{tab:res34} Test error rates (\%) of ResNet-34 and its AgrLearn counterparts on CIFAR-10, and CIFAR-100}
  \centering
  \begin{tabular}{c|cc|cc}
    \hline
    \multirow{2}{*}{\bf{Dataset}}  &\multicolumn{2}{c|}{\bf fold-1}       &\multicolumn{2}{c}{\bf fold-2}      \\  
       &$\alpha = 0$ &$\alpha = 0.7$    &$\alpha = 0$ &$\alpha = 0.3$       \\ \hline
    CIFAR-10          &  4.94  & 4.91   & 4.68  &   4.65   \\ 
    CIFAR-100        &	23.86   &    23.82  &    22.63   & 22.25 \\ \hline
    \end{tabular}
\end{table}

\subsubsection{Model Behavior During Training}
\begin{figure}[htb]
\centering
\includegraphics[width=0.9\columnwidth]{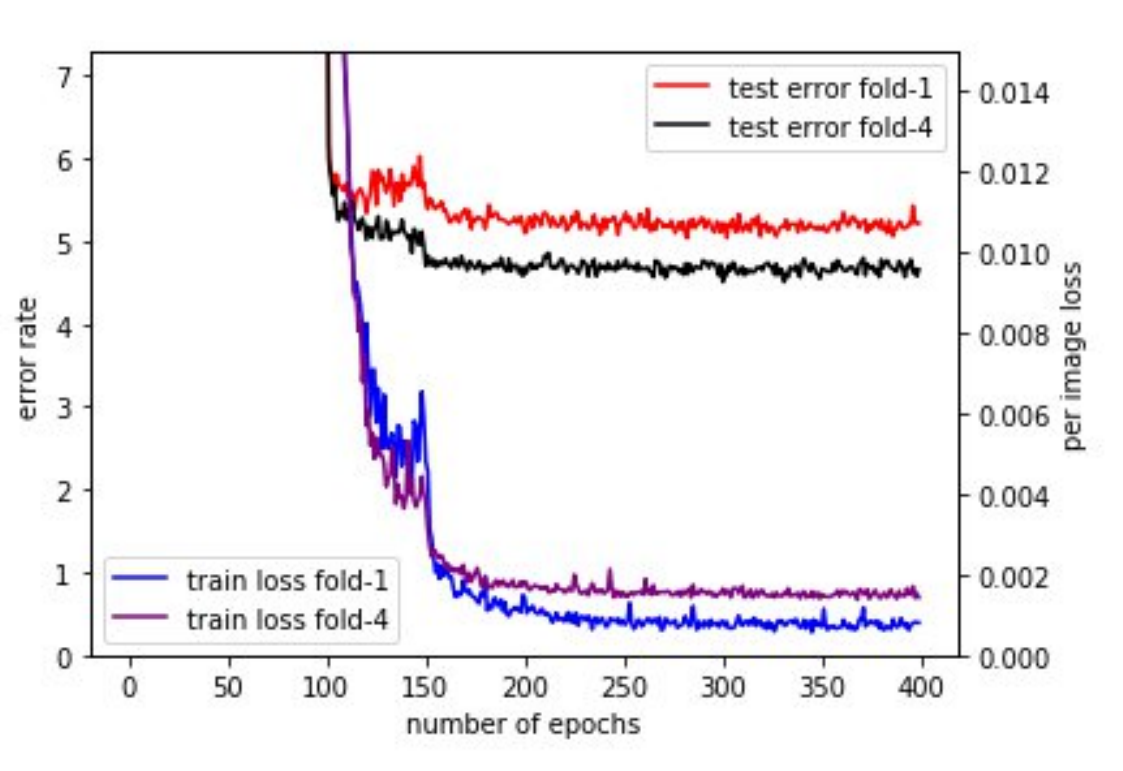}  
\caption{Training loss and test error on CIFAR-10.}
\label{fig3}
\end{figure}

The typical behavior of ResNet-18 for fold-1 and fold-4 (in terms of test error rate) across training epochs is shown in Figure \ref{fig3}. It is seen that in the ``stable phase" of training, the test error of fold-4  (black curve) continues to decrease whereas the test performance of fold-1 (red curve) fails to further improve. This can be explained by the training loss curve of fold-1 (blue curve), which drops to zero quickly in this phase and provides no training signal for further tuning the network parameters. In contrast, the training curve of fold-4 (purple curve) maintains a relatively high level, allowing the model to keep tuning itself. The relatively higher training loss of fold-4 is due to the much larger space of the amalgamated examples. Even in the stable phase, one expects that the model is still seeing new combinations of images. In other words, we argue that aggregating several examples into a single input can be seen as an implicit form of regularization, preventing the model from over-fitting by limited the number of individual examples.

\begin{table}[htb]
  \caption{\label{tab:dobla0} Test error rates (\%) of ResNet-18 (for fold-2, $\alpha = 0.3$) and its more complex variants} \centering
  \begin{tabular}{c|c|c}
  \hline
                                               &     CIFAR-10 & CIFAR-100      \\  \hline
            ResNet-18                  &  4.73     &  22.94     \\ 
      ResNet-18+double layer     &   4.3    &   21.78   \\ 
          ResNet-34                         &  4.65   &  22.25          \\ 
      ResNet-34+double layer    &  4.45    &  21.68   \\ \hline   
    \end{tabular}
 \end{table}

\subsubsection{Sensitivity to Model Complexity}
With fold-$n$ AgrLearn, the output label space becomes $\mathcal{Y}^n$. This significantly larger label space seems to suggest that AgrLearn favors a more complex model. In this study, we start with ResNet-18 for fold-2 and investigate the behavior of the model when it becomes more complex. The options we investigate include increasing the model width (by doubling the number of filters per layer) and increasing the model depth (from 18 layers to 34 layers). The performances of these models are given in Table \ref{tab:dobla0}.

Table \ref{tab:dobla0} shows that increasing the model width with respect to ResNet-18, and  ResNet-34, improves the performance of AgrLearn on both CIFAR-10 and CIFAR-100. For example, doubling the number of filters in ResNet-18 reduces the error rate for fold-2 where $\alpha$ is equal to $0.3$ from $4.73$\% to $4.3$\% on CIFAR-10, and from $22.94$\% to $21.78$\% on CIFAR-100, respectively. It also shows that increasing the model width with respect to ResNet-34 by factor 2, reduces the error rate from $4.65$\% to $4.45$\% on CIFAR-10, and from $22.25$\% to $21.68$\% on CIFAR-100. 

We hypothesize that with AgrLearn, the width of a model plays a critical role. This is because the input dimension in AgrLearn increases significantly and the model is required to extract joint features across individual objects in the amalgamated example. 

Moreover, increasing the model depth improves performance. For example, the relative error reductions from ResNet-18 to ResNet-34, where $\alpha$ is equal to $0.3$ are $1.7$\%, and $3$\% on CIFAR-10, and CIFAR-100 respectively. 

\subsubsection{Behavior with Respect to Fold Number}
We also conduct experiments investigating the performance of ResNet-18 with varying fold number $n$. Table \ref{tab:foldno} suggests that the performance of ResNet-18 is significantly boosted by increasing the number of folds $n$. For example, the relative error reductions achieved by fold-4,
where $\alpha$ is equal to $0$ are $4.72$\%, and $5.11$\% on CIFAR-10, and CIFAR-100, while the relative error reductions achieved by fold-2, are $3.74$\%, and $2.83$\% on CIFAR-10, and CIFAR-100. This shows that increasing the number of folds improves the performance of AgrLearn on both CIFAR-10 and CIFAR-100. Moreover, the relative error reductions achieved by fold-4, where $\alpha>0$ are $4.7$\%, and $5.8$\% on CIFAR-10, and CIFAR-100 respectively. 

\begin{table}[htb]
  \caption{\label{tab:foldno} Test error rates (\%) of ResNet-18 for varying fold numbers}
\resizebox{.98\columnwidth}{!}{ 
\begin{tabular}{c|cc|cc|cc}
    \hline
    \multirow{2}{*}{\bf{Dataset}}  &\multicolumn{2}{c|}{\bf fold-1}       &\multicolumn{2}{c|}{\bf fold-2}     &\multicolumn{2}{c}{\bf fold-4} \\  
                              &\bf{$\alpha = 0$}  &\bf{$\alpha = 0.7$}    &\bf{$\alpha = 0$} &\bf{$\alpha = 0.3$}       &\bf{$\alpha = 0$} &\bf{$\alpha = 4$} \\ \hline
    CIFAR-10          &   5.08      &  4.92     &  4.89      &  4.73       & 4.84  & 4.69  \\ 
    CIFAR-100        &	23.7    &   23.7    &  23.03     & 22.94 	    & 	22.49  &	22.32		\\ \hline
    \end{tabular} }
\end{table} 
\subsection{Text Classification}
 We test AgrLearn with two widely adopted NLP deep-learning architectures, CNN and LSTM~\cite{Hochreiter:1997:LSM:1246443.1246450}, using two benchmark sentence-classification datasets, Movie Review~\cite{pang2005seeing}
 %~\footnote{https://www.cs.cornell.edu/people/pabo/movie-review-data/} 
 and Subjectivity~\cite{DBLP:conf/acl/PangL04}. 
 Movie Review and Subjectivity contain respectively 10,662 and 10,000 sentences, with binary labels.  We use 10\% of random examples in each dataset for testing and the rest for training, as explained in~\cite{DBLP:conf/emnlp/Kim14}. 
 
For CNN, we adopt CNN-sentence~\cite{DBLP:conf/emnlp/Kim14} and implement it exactly as~\cite{kim/code}. For LSTM, we just simply replace the convolution and pooling components in CNN-sentence with standard LSTM units as implemented in~\cite{Abadi:2016:TSL:3026877.3026899}. The final feature map of CNN and the final state of LSTM are passed to a logistic regression classifier for label prediction. 
Each sentence enters the models via a learnable, randomly initialized word-embedding dictionary. For CNN, all sentences are zero-padded to the same length.

\begin{table}[htb]
  \caption{\label{tab:accuracy:cnn} Accuracy (\%) obtained by CNN, LSTM and their respective AgrLearn models}
  \centering
  \begin{tabular}{c|cc|cc}
    \hline
    \multirow{2}{*}{\bf{Dataset}}  &\multicolumn{2}{c|}{\bf CNN}       &\multicolumn{2}{c}{\bf LSTM}      \\  
                              & fold-1 &fold-2   & fold-1 & fold-2       \\ \hline
    Movie Review    &76.1   & 79.3 & 76.2 & 77.8\\ 
    Subjectivity      &90.01 & 93.5 & 90.2 & 92.1 \\ \hline
    \end{tabular}
\end{table}

The fold-2 AgrLearn model corresponding to the CNN and LSTM models are constructed, where $\alpha$ is equal to $0$. In CNN with fold-2, the aggregation of two sentences in each input simply involves concatenating the two zero-padded sentences. In LSTM with fold-2, when two sentences are concatenated in tandem, an EOS word is inserted after the first sentence. 

We train and test the CNN, LSTM and their respective AgrLearn models on the two datasets, and report their performance in Table~\ref{tab:accuracy:cnn}. Clearly, the AgrLearn models improve upon their corresponding CNN or LSTM counterparts. In particular, the relative performance gain brought by AgrLearn on the CNN model appears more significant, amounting to $4.2$\% on Movie Review and $3.8$\% on Subjectivity.

\section{Conclusion}

Aggregated Learning, or AgrLearn,  is a simple and effective neural network modeling framework, justified information theoretically. It builds on an equivalence between IB learning and IB quantization and exploits the power of vector quantization, which is well known in information theory. We have demonstrated its effectiveness through the significant performance gain it brings to the current art of deep network models. 

We believe that the proposal and successful application of AgrLearn in this paper signals the beginning of a promising and rich theme of research. Many interesting questions deserve further investigation. For example, how can we characterize the interaction between model complexity, fold number and sample size in AgrLearn? Additionally, the aggregation of inputs 
provides additional freedom in the architectural design of the network; how can such freedom be better exploited?

\section{Acknowledgments}

This work is supported partly by the National Natural Science Foundation of China (No. 61772059, 61421003), by the Beijing Advanced Innovation Center for Big Data and Brain Computing (BDBC).

%%%=====================================

\section{Appendices}
Here we give a brief review of typical sequences~\cite{orlitsky2001coding}, which will be useful in proving Theorem 1. We remark that the notion of typicality here is stronger than the widely
used (weak) typicality in \cite{cover2006wiley}, and refer the interested reader
to~\cite{el2011network} for a comprehensive treatment of the subject. 
Throughout this note, the symbol $\mathbb{E}$ will denote expectation. At some places, we might use
subscripts to explicitly indicate the random variables with respect to which the expectation is
performed.  

\begin{enumerate}
\item 
\underline{Empirical distribution:}
Given a sequence $x^n \in \mathcal{X}^{n}$, it induces an empirical distribution on $\mathcal{X}$ defined as
\begin{equation}
\pi(x|x^n) := \frac{1}{n}|\{i:x_i = x\}| \text{ for all } x \in \mathcal{X}
\end{equation}
\item
\underline{Typical set:}
For $X\sim p_{X}(x)$ and  $\epsilon \in (0,1)$, the set of $\epsilon$-typical sequences is defined as
\begin{align}
\begin{split}
\mathcal{S}_{\epsilon}^{n}(X) :=& \{ x^{n} \vert |\pi(x|x^n) - p_X(x)| \leq \epsilon p_X(x)\\
&\text{ for all } x\in \mathcal{X} \}
\end{split}
 \end{align}

\item 
\underline{Typical average lemma:} For any $x^n \in \mathcal{S}_{\epsilon}^{n}(X)$ and any non-negative function $g$ on $\mathcal{X}$, we have 
\begin{equation}
(1-\epsilon) \mathbb{E}[g(X)] \leq \frac{1}{n} \sum_{i}g(x_i) \leq (1+\epsilon) \mathbb{E}[g(X)]
\end{equation}

Note that by choosing $g$ to be the $\log$ function, one recovers the notion of typicality in
\cite{cover2006wiley}.
The typicality here is strictly stronger than the one in \cite{cover2006wiley}, however,
similar to weak typicality, most i.i.d. sequences are still typical under this definition. Namely, for any i.i.d sequence $X^n$ of RVs with $X_i \sim p_{X}(x_i)$, by the LLN,
the empirical distribution $\pi(x|X^n)$ converges (in probability) to $p_{X}(x)$, for all
$x\in \mathcal{X}$, and so such sequence, with high probability, belongs to the typical set.
\item
\underline{Joint typicality:} Items~1 and 2 extend to a joint source $(X,Y) \sim
p_{XY}(x,y)$ in the obvious way, i.e., by treating $X$ and $Y$ as one source $(X,Y)$.
Given a sequence $(x^n, y^{n}) \in \mathcal{X}^{n}\times \mathcal{Y}^{n}$, it induces an empirical distribution on $\mathcal{X} \times \mathcal{Y}$ defined as

\begin{align}
\begin{split}
\pi(x,y|x^n,y^n) :=& \frac{1}{n}|\{i:x_i = x, y_i = y\}| \\
& \text{ for all } (x,y) \in \mathcal{X} \times \mathcal{Y}
\end{split}
 \end{align}

For $X\sim p_{X}(x)$ and  $\epsilon \in (0,1)$, the set of $\epsilon$-typical sequences is defined as

\begin{align}
\begin{split}
\mathcal{S}_{\epsilon}^{n}(X,Y) :=& \{ (x^{n},y^n) \vert |\pi(x,y|x^n,y^n) - p_{XY}(x,y)|\\
& \leq \epsilon p_{XY}(x,y) \text{ for all } (x,y)\in \mathcal{X} \times \mathcal{Y} \}
\end{split}
 \end{align}

\item 
\underline{Joint typicality lemma:} Let $(X,Y) \sim p_{XY}(x,y)$ and $p_{Y}(y)$ be the
marginal distribution $\sum_{x}p_{XY}(x,y)$. Then, for $\epsilon' < \epsilon$,
there exists $\delta(\epsilon) \rightarrow 0$ as $\epsilon \rightarrow 0$ such that
\begin{equation}
p \{(x^n,Y^n) \in \mathcal{S}_{\epsilon}^{n}(X,Y) \} \geq 2^{-n(I(X;Y)+\delta(\epsilon))}
\label{eq:jt-lemma}
\end{equation}
for $x^n\in \mathcal{S}_{\epsilon'}^n$, $Y^n\sim \prod_{i = 1}^n p_{Y}(y_i)$, and
sufficiently large $n$.
\end{enumerate}
%%%%%%%%%%%%%%%%%%%%%%%%%%%%%%%%%%%%%%%%%%%%%%%%
\subsection{Proof of Theorem 1}
We should make a few remarks before presenting a proof.
The proof follows standard techniques from information theory for proving results of this nature. 
It is worth noting that the conventional proof of achievability \cite{cover2006wiley} of the rate-distortion theorem does not directly apply here since the distortion measure $d_{\rm IB}$ depends on the distribution $p_{T|X}$. This was addressed in~\cite{gilad2003information} by extending the definition of distortion jointly typical sequences in~\cite{cover2006wiley} to multi-distortion jointly
typical sequences. Our approach exploits the notion of typicality presented in the previous section
and closely follows the proof of achievability in~\cite{el2011network} of the rate-distortion theorem.
\begin{equation}
\label{eq:rate-info}
R'(D) := \min_{p_{T|X}(t|x): \mathbb{E}[d(X,T)] \leq D} I(X;T)
\end{equation}
We need to show $R_{\rm IBQ}(D) = R'(D)$.

\subsubsection{Proof of the converse:}
We first show $R_{\rm IBQ}(D) \geq R'(D)$ by showing that for any sequence of $(n,2^{nR})$ codes satisfying $\mathbb{E}\overline{d}_{\rm IB}(X^n,T^n)\leq D$, it must be the case that $R \geq R'(D)$.
We have
\begin{align}
\begin{split}
nR
&\overset{\text{(i)}}{\geq} H(f_{n}(X^{n}))
\overset{\text{(ii)}}{\geq} I(X^{n};f_{n}(X^n))
\overset{\text{(iii)}}{\geq} I(X^{n},T^{n}) \\
&= \sum_{i}H(X_i) - H(T_i | X^{n}, T^{i-1}) \\
& \geq \sum_{i}H(X_i) - H(T_i|X_i) = \sum_{i} I(X_i; T_{i}) \\
&\overset{\text{(iv)}}{\geq} \sum_{i} R'(\mathbb{E}[d(X_{i},T_i)]) \overset{\text{(v)}}{\geq} n  R'(\frac{1}{n}\sum_{i}\mathbb{E}[d(X_{i},T_i)]) \\
&\overset{\text{(vi)}}{=} n  R'(\mathbb{E}[d(X^{n},T^n)]) \overset{\text{(vii)}}{\geq} n  R'(D)
\end{split}
 \end{align}
%\stackrel
where
(i) follows from the fact that $f_{n}$ takes its values from $\{1,\dots, 2^{n}\}$,
(ii) from the non-negativity of conditional entropy,
(iii) from the data processing inequality since $T^n = g_n(f_n(X^n))$,
(iv) from  (\ref{eq:rate-info}) by noting that $R'(\mathbb{E}[d(X_{i},T_i)]) = \min_{p_{T_i|X_i}} I(X_i;T_i)$,
(vi) by definition of $\overline{d}_{\rm IB}$ and
(vii) from $\mathbb{E}\overline{d}_{\rm IB}(X^n,T^n)\leq D$ since $R'(D)$ is a decreasing function in $D$.
To prove (v), it is sufficient to show that $R'$ is a convex function in $D$, which is shown in the following lemma.
%\qed

\begin{lem}
\cite{ahlswede1975source}. The function $R'(D)$ defined in (\ref{eq:rate-info}) is a convex function.
\label{lem1}
\end{lem}
Proof.
Let $(D_1,R_1)$ and $(D_2,R_2)$ be two points on $R'(D)$ attained, respectively, by 
$T_1$ and $T_2$ via the minimizers $p_{T_1|X}$ and $p_{T_2|X}$ of
(\ref{eq:rate-info}).
Define 

\begin{align}
\begin{split}
T = \left\{
\begin{array}{ll}
T_1, & Z = 1 \\
T_2, & Z = 2
\end{array}
\right.
\end{split}
\end{align}

where $Z\in \{1,2\}$ is a RV independent of $(T_1,T_2,X,Y)$ with $p_{Z}(1) = \lambda$.
Then,

\begin{align}
\begin{split}
p_{XTZ}(x,t,z) = \left\{
\begin{array}{ll}
\lambda \cdot  p_{XT_{1}}(x,t), & Z = 1 \\
(1-\lambda) \cdot p_{XT_{2}}(x,t), & Z = 2
\end{array}
\right.
\end{split}
\end{align}

and so

\begin{align}
\begin{split}
I(X;T,Z) &= \sum_{x,t,z}
p_{XTZ}(x,t,z)\log\frac{p_{XTZ}(x,t,z)}{p_{X}(x)p_{TZ}(t,z)}  \\
& = \sum_{x,t} \lambda \cdot p_{XT_1}(x,t)\log \frac{\lambda \cdot
p_{XT_1}(x,t)}{\lambda \cdot p_{X}(x)p_{T_1}(t)} \\
&+   \sum_{x,t} (1-\lambda)\cdot p_{XT_2}(x,t)\\
&\times \log \frac{(1-\lambda)\cdot
p_{XT_2}(x,t)}{(1-\lambda)\cdot p_{X}(x)p_{T_2}(t)}
\\ & = 
\lambda \cdot I(X;T_1) + (1-\lambda) \cdot  I(X;T_2)
\end{split}
\end{align}

Moreover, we have

\begin{align}
\begin{split}
\mathbb{E}[d(X,(T,Z))]
& = \sum_{x,t,z}p_{X T Z}(x,t,z)\\
& \times \sum_{y}p_{Y|X}(y|x)\log \frac{p_{Y|X}(y|X)}{p_{Y|TZ}(y|t,z)} \\
& = H(Y|TZ) - H(Y|X) \\
& = \lambda \cdot H(Y|T_1) + (1-\lambda)\cdot H(Y|T_2) \\
& - \lambda\cdot H(Y|X) - (1-\lambda)\cdot H(Y|X) \\
& = \lambda\cdot \mathbb{E}[d(X,T_1)] + (1-\lambda)\cdot \mathbb{E}[d(X,T_2)]
\end{split}
\end{align}

Since $(T,Z)$---$X$---$Y$ is a markov chain resulting in cost and constraint that are
linear functions of the original costs and constraints, the claim follows from the definition of
$R'$ in (\ref{eq:rate-info}).

%%%%%%%%%%%%%%%%%%%%%%%%%%%%%%%%%%%%%%%%%%%%%%%%

\subsubsection{Proof of Achievability in Theorem 1:}
We need to show that for $R = R'(D)$ there exists a sequence $(2^{nR},n)$ of codes satisfying
$\mathbb{E}\overline{d}_{\rm IB}(X^n,T^n)\leq D$ .

\underline{Random codebook:}
Let $R = R'(D)$ and fix $p_{T|X}$ to be an optimal distribution to the minimization
(\ref{eq:rate-info}) at $D/(1+\epsilon)$, i.e., we pick a conditional distribution that attains
$R'(D/(1+\epsilon))$.
%%%%%%
\footnote{A comment on existence. There is a feasible distribution $p_{T|X}$ satisfying the distortion constraint for any $D$. For $D = 0$, choose $p_{T|X}(t|x) = p_{X}(t)$ and for $D\geq D_{\max}:=I(X;Y)$ choose $p_{T|X}$ as the degenerate distribution that assigns all the weight on one element of $T$. For $D \in [0,D_{\max}]$, use a latent variable $Z$ as in the proof of the Lemma \ref{lem1} with $\lambda = D/D_{\max}$.}
%%%%%%
Let $p_{T}(t) = \sum_{x\in \mathcal{X}} p_{X}(x) p_{T|X}(t|x)$. Generate $2^{nR}$ i.i.d.
sequences $t^{n}(m) \sim \prod_{i=1}^{n} p_{T}(t_i)$, $m\in \{1,\dots, 2^{nR}\}$.
These sequences form the codebook which is revealed to the encoder and decoder.

\underline{Encoder:} The encoder uses joint typicality encoding. Given a sequence $x^{n}$, find an
index $m$ s.t. $(x^{n},t^{n}(m)) \in \mathcal{S}_{\epsilon}^{n}(X,T)$ and send $m$. If there is more than one index then choose $m$ to be the smallest index, and if there is no index then choose $m = 1$. 
(In other words, the encoder sets $f_n(x^n)$ to be the index $m$, where $m$ is as described above.)

\underline{Decoder:} Upon receiving index $m$, set $t^n = t^n(m)$.
(In other words, the decoder sets $g_n(m)$ to be the row of the codebook indexed by $m$.)

\underline{Expected distortion}
Let $\epsilon' < \epsilon$ and $M$ be the index chosen by the encoder. We first bound the distortion
averaged over codebooks. Towards this end, define the event

\begin{equation}
\mathcal{E} := \{ (X^n,T^n(m)) \notin \mathcal{S}_{\epsilon}^{n}(X,T) \}
\end{equation}
then by the union bound and the choice of the encoder, we have
\begin{equation}
p(\mathcal{E}) \leq p(\mathcal{E}_1) + p(\mathcal{E}_2)
\end{equation}
where

\begin{align}
\begin{split}
\mathcal{E}_{1} &:= \{ X^{n} \notin \mathcal{S}^n_{\epsilon'}(X) \}, \\
\mathcal{E}_{2} &:= \{ X^{n} \in \mathcal{S}^n_{\epsilon'}, (X^n,T^n(m)) \notin \mathcal{S}_{\epsilon}^n(X,T)\\
&  \forall m \in \{1,\dots, 2^{nR}\} \} 
\end{split}
\end{align}

We have $\lim_{n\rightarrow \infty} p(\mathcal{E}_{1}) = 0$ by the LLN and 
\begin{align}
\begin{split}
p(\mathcal{E}_{2})
&= \sum_{x^{n}\in \mathcal{S}_{\epsilon'}^{n}} p_{X^n}(x^n) \\
&\times p\big\{(x^n,T^n(m) ) \notin \mathcal{S}_{\epsilon}^{n}  \forall m \mid X^{n} = x^{n}) \big\} \\
&\stackrel{\text{(i)}}{=}
\sum_{x^{n}\in \mathcal{S}_{\epsilon'}^{n}} p_{X^n}(x^n)
\prod_{m=1}^{2^{nR}} p\big\{(x^n,T^{n}(m)) \notin \mathcal{S}_{\epsilon}^{n} \big\} \\
&\stackrel{\text{(ii)}}{=}
\sum_{x^{n}\in \mathcal{S}_{\epsilon'}^{n}} p_{X^n}(x^n)
\big(p\big\{(x^n,T^{n}(1)) \notin \mathcal{S}_{\epsilon}^{n} \big\} \big)^{2^{nR}} \\
&\stackrel{\text{(iii)}}{\leq}
\sum_{x^{n}\in \mathcal{S}-{\epsilon'}^{n}} p_{X^n}(x^n)
\big(1- 2^{-nI(X;T)+\delta(\epsilon))} \big)^{2^{nR}} \\
&\stackrel{\text{}}{\leq}
\big(1- 2^{-nI(X;T)+\delta(\epsilon))} \big)^{2^{nR}} \\
&\stackrel{\text{(iv)}}{\leq}
\exp\big(-2^{n(R-I(X;T) - \delta(\epsilon))}\big)
\end{split}
\end{align}

where (i) and (ii) are by the i.i.d assumption on the codewords, (iii) is by the joint typicality
lemma, (iv) is by the fact $(1-\alpha)^{k} \leq \exp(-k\alpha)$ for $\alpha \in [0,1]$ and $k\geq 0$.
Hence, we have $\lim_{n\rightarrow \infty} p(\mathcal{E}_2) = 0$ for $R > I(X;T)+\delta(\epsilon)$.

Now, the distortion averaged over $X^{n}$ and over the random choice of the codebook is given as

\begin{align}
\begin{split}
&\mathbb{E}_{X^{n},T^{n},M} [d(X^{n},T^{n}(M))]\\
&=
p(\mathcal{E}) \cdot \mathbb{E}_{X^{n},T^{n},M} [d(X^{n},T^{n}(M)) | \mathcal{E}]\\
&+ p(\mathcal{E}^{c}) \cdot \mathbb{E}_{X^{n},T^{n},M} [d(X^{n},T^{n}(M)) | \mathcal{E}^c] 
\\ 
&\leq p(\mathcal{E}) \cdot d_{\max}
+ p(\mathcal{E}^{c}) \cdot \mathbb{E}_{X^{n},T^{n},M} [d(X^{n},T^{n}(M)) | \mathcal{E}^c]
\\
&= p(\mathcal{E}) \cdot d_{\max}
+ p(\mathcal{E}^{c}) \cdot \mathbb{E}_{X^{n},T^{n}} [d(X^{n},T^{n}(1)) | \mathcal{E}^c]
\\
&\leq p(\mathcal{E}) \cdot d_{\max}
+ p(\mathcal{E}^{c})\cdot (1+\epsilon)\cdot \mathbb{E}_{X,T} [d(X,T)]
\end{split}
\end{align}

where $d_{\max} = \max_{(x,t)\in\mathcal{X} \times \mathcal{T}}d(x,t)$.
By the choice of $p_{T|X}(t|x)$, we have $\mathbb{E}[d(X,T)]\leq D/(1+\epsilon)$, and so
\begin{equation}
\lim_{n\rightarrow \infty} \mathbb{E}_{X^{n},T^{n},M}[d(X^{n},T^n(M))] \leq D
\end{equation}
for $R> I(X,T) + \delta(\epsilon)$, where $\delta(\epsilon) \rightarrow 0$ as $n\rightarrow
\infty$.
Since the expected distortion, averaged over codebooks, satisfies the distortion constraint $D$, there
must exist a sequence of codes that satisfies the constraint. This shows the achievability of the
rate-distortion pair $(R(D/(1+\epsilon)+\delta(\epsilon), D)$. 
By the continuity of $R(D)$ in $D$ the achievable rate $R(D/(1+\epsilon)) + \delta(\epsilon)$
converges to $R(D)$ as $\epsilon \rightarrow 0$.
%\qed

%\subsection{Proof of Theorem 2}
%Follows directly from Theorem 1 and the definition of $d_{\rm IB}$. Namely, we have

%\begin{align}
%\begin{split}
%\mathbb{E}_{X,T}(d_{\rm IB}(X,T)) :=&\sum_{x,t} d_{\rm IB}(x,t) p_{XT}(x, t)  \\
%=& I(X;Y) - I( T ;Y)
%\label{eq:dist-expect3}
%\end{split}
%\end{align}

%where the second equality is by the definition of $d_{\rm IB}$ and the Markov chain $T\text{---}X\text{---}Y$.
%Hence, we have

%\begin{align}
%\begin{split}
%R_{\rm IBQ}(D) = & \min_{p_{T|X}(t|x): \mathbb{E}[d(X,T)] \leq D} I(X;T) \\
%= & \min_{p_{T|X}(t|x): I(Y;T) \geq I(X;Y) -  D} I(X;T) \\
%= & R_{\rm IBL}(I(X;Y) - D)
%\end{split}
%\end{align}

%where the first equality is by Theorem 1, the second by (\ref{eq:dist-expect3}), and the third by the definition of $R_{\rm IBL}$. The claim follows via the substitution $A:=I(X;Y) - D$.

\subsection{Proof of Theorem 3}
\begin{equation}
    \setlength{\jot}{10pt}
    \begin{aligned}[b]
 I(Y;T) 
 & = \sum_{(t,y)\in \mathcal{T}\times \mathcal{Y}}  p_{YT}(y, t) \log\frac{p_{Y|T}(y|t)}{p_Y(y)} \\
& =  \sum_{(t,y)\in \mathcal{T}\times \mathcal{Y}}   p_{YT}(y, t) \log
\frac{p_{Y|T}(y|t)}{q_{Y|T}(y|t)} 
\cdot \frac{q_{Y|T}(y|t)}{p_Y(y)}\\
& =  \sum_{(t,y)\in \mathcal{T}\times \mathcal{Y}}   p_{YT}(y, t) \log q_{Y|T}(y|t) \\
&+ \sum_{(t,y)\in \mathcal{T}\times \mathcal{Y}}    p_{YT}(y, t)\log \frac{p_{Y|T}(y|t)}{q_{Y|T}(y|t)} \\
& -  \sum_{(t,y)\in \mathcal{T}\times \mathcal{Y}}   p_{YT}(y, t) \log p_Y(y)\\ 
& =  
%\mathbb{E}_{t\sim p_T}  \left\{-{\rm CE}\left(p_{Y|T}(\cdot|t)\Vert q_{Y|T}(\cdot|t)\right)\right\}
\sum_{(t,y)\in \mathcal{T}\times \mathcal{Y}}    p_{YT}(y, t) \log q_{Y|T}(y|t) \\
&+ \mathbb{E}_{t\sim p_T(t)}  \text{KL}\left(p_{Y|T}(\cdot|t)\Vert q_{Y|T}(\cdot|t)\right) +H(Y)\\
 & \ge   
\sum_{(t,y)\in \mathcal{T}\times \mathcal{Y}}    p_{YT}(y, t) \log q_{Y|T}(y|t)
  +H(Y)\\
& =  
\mathbb{E}_{(x,y)\sim p_{XY}(x,y), \atop{ t\sim p_{T|X}(.|x)}}  \log q_{Y|T}(y|t)
  +H(Y)
\label{eq:var}
 \end{aligned}
\end{equation}
Note that the inequality above is due to the non-negativity of KL-divergence, in which equality is achieved precisely when $q_{Y|T}(y|t)$ is identically equal to $p_{Y|T}(y|t)$.

\bibliographystyle{aaai}
\bibliography{aaai}

\end{document}